\documentclass[10pt,twocolumn,letterpaper]{article}

\usepackage{cvpr}
\usepackage{times}
\usepackage{epsfig}
\usepackage{graphicx}
\usepackage{amsmath}
\usepackage{amssymb}
\usepackage{multirow}
\usepackage{dsfont}
\usepackage{subfiles}
\usepackage{subcaption}
\usepackage{authblk}

\usepackage[pagebackref=true,breaklinks=true,letterpaper=true,colorlinks,bookmarks=false]{hyperref}

\usepackage[breaklinks=true,bookmarks=false]{hyperref}

\cvprfinalcopy

\begin{document}

\title{\vspace{-1.0cm}Visualization, Discriminability and Applications of Interpretable Saak Features}

\author{Abinaya Manimaran$\dagger$, Thiyagarajan Ramanathan$\dagger$, Suya You$\ddagger$, C-C Jay Kuo$\dagger$}
\affil{$\dagger$University of Southern California, Los Angeles, California, USA \\
 $\ddagger$US Army Research Laboratory, Playa Vista, California, USA}

\maketitle

\begin{abstract}

In this work, we study the power of Saak features as an effort towards interpretable deep learning.  Being inspired by the operations of convolutional layers of convolutional neural networks, multi-stage Saak transform was proposed. Based on this foundation, we provide an in-depth examination on Saak features, which are coefficients of the Saak transform, by analyzing their properties through visualization and demonstrating their applications in image classification. Being similar to CNN features, Saak features at later stages have larger receptive fields, yet they are obtained in a one-pass feedforward manner without backpropagation. The whole feature extraction process is transparent and is of extremely low complexity. The discriminant power of Saak features is
demonstrated, and their classification performance in three well-known datasets (namely, MNIST, CIFAR-10 and STL-10) is shown by experimental results. 

\end{abstract}

\vspace*{-5mm}
\section{Introduction} \label{sec:Introduction}

The quality of image features is crucial for a wide range of image understanding and computer vision tasks including object detection, segmentation, classification, and recognition. All these higher-level tasks have traditionally relied on handcrafted features by feature engineering that intends to capture the essence of different visual patterns. Much recent work has been focused on automatically learning good feature representations from a massive amount of input data. Nevertheless, both feature engineering and feature learning have their advantages in feature representations, which remains active research in computer vision and machine learning domains. Good feature representations typically should be discriminative, robust, concise, and computationally effective \cite{Boureau10learningmid-level, Lazebnik, Ref1}. 

Before the surge of convolutional neural networks (CNNs), feature extraction was most often conducted in an unsupervised manner. That is, feature extraction and the decision-making (e.g. classifier or detector) modules are completely decoupled. Such framework however has been changed in the feature learning architecture. The CNN solution has been widely used in computer vision and image processing tasks nowadays. One reason of CNN's success lies in the end-to-end system optimization where there exists no clear boundary between the feature extraction and the decision-making modules. Thus, CNN features are label-dependent (i.e., supervised training) and obtained through backpropagation. The strong coupling between feature extraction and decision-making makes the whole CNN mechanism difficult to explain. Another striking property of CNNs is that the multi-layer convolutional
layers can extract features of a large spatial size, which is clearly shown through the visualization study in \cite{VisualizationZeiler}.

To explain the superior performance of CNNs, Kuo {\em et al.} published a sequence of papers on interpretable CNNs 
\cite{DBLP:journals/corr/Kuo16, DBLP:journals/corr/Kuo17}. Based on the idea of ``Subspace approximation with augmented kernels", Kuo and Chen proposed a new transform called
the Saak transform in \cite{DBLP:journals/corr/abs-1710-04176}. The Saak transform is a variant of principal component analysis (PCA) that splits the positive and negative response outputs into two separate channels through kernel augmentation. The kernel augmentation process facilitates the cascade of multi-stage Saak transforms by resolving "sign confusion" ambiguity. Being similar to CNN features,
Saak features at later stages have larger receptive fields, yet they are obtained in a one-pass feedforward manner without any supervision and backpropagation. The whole feature extraction process is completely transparent and of extremely low complexity. In addition, Saak transform is invertible, allowing the Saak feature representations to be transformed back to the image space for clearly visualizing, analyzing, and interpreting. 

The Saak coefficients were adopted as features in classifying hand-written digits under various noisy environments in \cite{DBLP:journals/corr/abs-1710-10714}. The Saak-based solution outperforms those by the standard CNN in term of classification accuracy in most cases. Furthermore, Saak transform can be used as a pre-processing step and applied to adversarial images. The impact of these attacks can be mitigated significantly \cite{song2018defense}. 

Despite of the above-stated work, research on multi-stage Saak transforms and the corresponding features is relatively
less. In this work, we provide an in-depth examination on the unique properties of Saak features in terms of discriminant power, robustness, and complexity. Our contributions are as follows. First, we have several generalizations on top of the basic Saak transform. Multiple Saak transforms are cascaded to transform images of a larger size from the spatial domain to a joint spatial-spectral domain. Second, we develop new strategies for Saak feature selections and analyze their properties through visualization to get insight into the Saak functions at different stages. Third, we apply the new Saak feature representations to the problems of image classification and defense against adversarial attacks. Their performances with three well-known benchmark datasets (namely, MNIST, CIFAR-10 and STL-10) are demonstrated by experimental results.

%%%%%%%%%%%%%%%%%%%%%%%%%%%%%%%%%%%%%%%%%%%%%%%%%
\begin{figure}
\begin{center}
\includegraphics[width=1\linewidth]{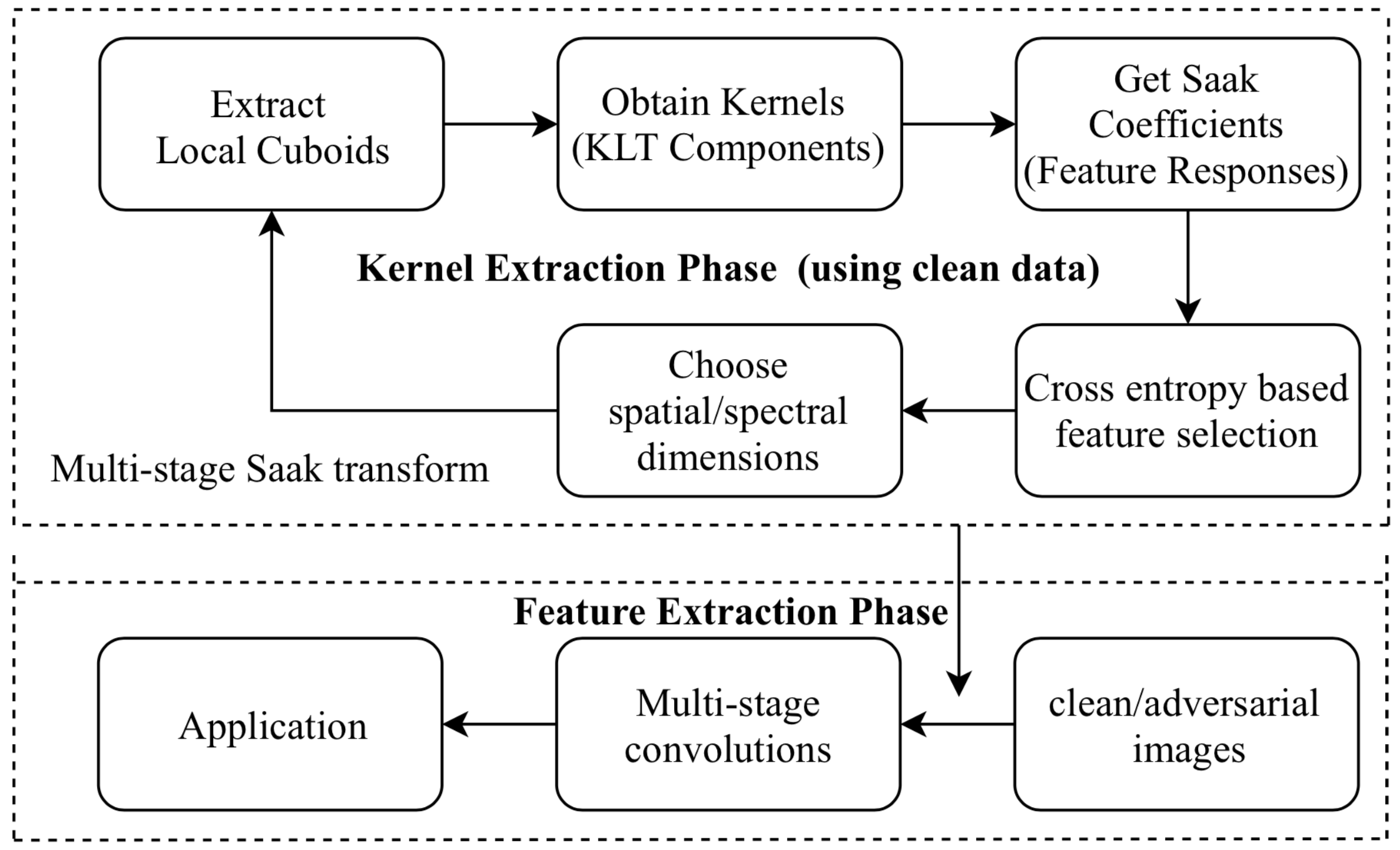}
\end{center}
\vspace*{-5mm}
\caption{Saak transform consists two modules: kernel extraction and feature extraction. We use training images
to extract kernels followed feature extraction. Feature responses can used for any application.}
\label{fig:flow chart}
\end{figure}
%%%%%%%%%%%%%%%%%%%%%%%%%%%%%%%%%%%%%%%%%%%%%%%%%

%%%%%%%%%%%%%%%%%%%%%%%%%%%%%%%%%%%%%%%%%%%%%%%%%
\begin{figure*}
\begin{center}
\includegraphics[width=1\linewidth]{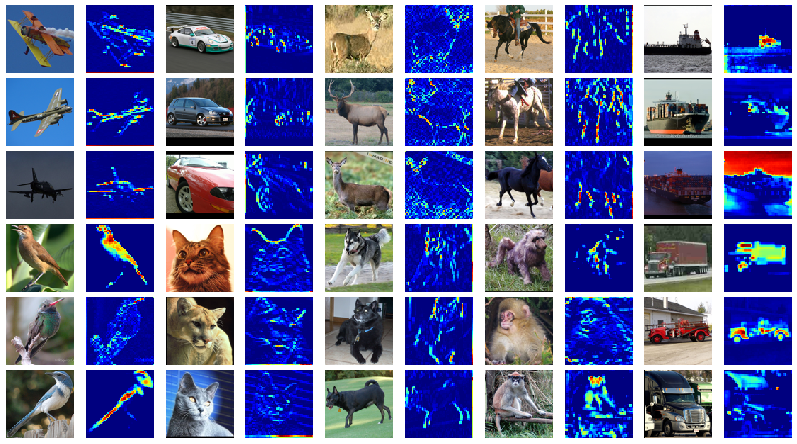}
\end{center}
\vspace*{-5mm}
\caption{Visualization of feature responses from STL-10 (three images from each class).
Each pair contains the original image and it's first stage Saak
response. Discriminant feature for every class is highlighted.}
\label{fig:Figure-1}
\end{figure*}
%%%%%%%%%%%%%%%%%%%%%%%%%%%%%%%%%%%%%%%%%%%%%%%%%
\vspace*{-0.7mm}
\section{Saak Transform and Generalizations} \label{sec:Saak transform}

The Saak transform defines a mapping from a real-valued function defined on a three-dimensional (3D) cuboid consisting of spatial and spectral dimensions to a one-dimensional (1D) rectified spectral vector. It is presented as a new feature representation method. It consists of two main ideas: subspace approximation and kernel augmentation. For the former, we build the optimal linear subspace approximation to the original signal space via PCA or the truncated Karhunen-Loève Transform (KLT) \cite{Stark:1986:PRP:5639}. For the latter, we augment each transform kernel with its negative and apply the rectified linear unit (ReLU) to the transform output. This is equivalent to the sign-to-position (S/P) format conversion. 

We review the Saak transform briefly in this section. For input samples {\boldmath$\tilde{f}$} $\in$ $R^N$, we can obtain its DC component
by projecting {\boldmath$\tilde{f}$} onto the dc (direct current) unit vector in form of,  
\begin{equation} \label{eqn:DC}
{\bf b}_0 = \frac{1}{\sqrt{N}}{(1,1,\cdots,1)^T} = {\bf a}_0.
\end{equation}

The dc-removed samples contain only ac (alternating current) 
components. They can be written as,
\begin{equation} \label{eqn:AC}
{\bf f} = \tilde{\bf f} - {\bf b}_0^T \tilde{\bf f}.
\end{equation}
The correlation matrix of ${\bf f}$ is 
\begin{equation}
R = E[{\bf f} \; {\bf f}^T] \in R^{N \times N},
\end{equation}
and its eigenvectors are orthogonal vectors:
\begin{equation} \label{eqn:orthonormal}
{\bf b}_i^T {\bf b}_j = <{\bf b}_i, {\bf b}_j> = \delta_{i,j}.
\end{equation}
They are kernels of the one-stage Saak transform.  By kernel augmentation, we choose both ${\bf b}$ and $-{\bf b}$ as new kernels.
That is, we define
\begin{equation} \label{eqn:augmentedkernels}
{\bf a}_{2k-1} = {\bf b}_k, \hspace{3mm} {\bf a}_{2k} = - {\bf b}_k; 
\hspace{3mm} k=1,2,\cdots, N-1
\end{equation}

Projection of input \textbf{f} onto augmented kernels yields Saak  coefficients in form of
\begin{equation} \label{eqn:projection}
p_k = {\bf a}_k^T\textbf{f}, \quad k=0, 1, \cdots, 2N-1,
\end{equation}
The application of the ReLU activation to the projected vector (except for the DC component) gives the following output,
\begin{equation}\label{eqn:relu}
\textbf{g} = (g_0, g_1,.....,g_{2N-1})^T \hspace{1mm} \in R^{2N-1} \\
\end{equation}
where $g_0 = p_0$ and
\begin{itemize}
\item If $p_{2k-1} \ge 0$, $g_{2k-1} = p_{2k-1}$ and $g_{2k} = 0$
\item If $p_{2k} \ge 0$, $g_{2k-1} = 0$ and $g_{2k} = p_{2k}$.
\end{itemize}
Thus, Saak coefficients are always non-negative.

Multi-stage Saak transforms are developed to transform images of large spatial dimensions. The spectral dimension of the lossless Saak transform \cite{DBLP:journals/corr/abs-1710-04176} grows very rapidly as the stage number becomes large. The lossy Saak transform \cite{DBLP:journals/corr/abs-1710-10714} adopts a truncated KLT. 

We introduce several generalizations to the basic Saak transform. First, Saak transform in \cite{DBLP:journals/corr/abs-1710-04176} used non-overlapping windows. Instead, we adopt overlapping windows with stride equal to one. This is because we do not know the optimal spatial location for pattern matching and the redundant representation offers a richer feature set to select in the later decision-making module.  Second, the F-test was applied as a criterion for feature selection in \cite{DBLP:journals/corr/abs-1710-04176} while a cross-entropy-based metric is adopted for feature selection in Section
\ref{subsec:Feature_Selection}. Third, the spatial dimension of all transform kernels is set to $2 \times 2$ in
\cite{DBLP:journals/corr/abs-1710-04176}. We conduct an extensive evaluation of kernels of different spatial sizes in this work.  For the MNIST dataset, a kernel of spatial dimension $2 \times 2$ appears to be sufficient since the variation within an image is relatively simple. However, the CIFAR-10 and the STL-10 datasets contain more complex objects with diversified background. Kernels of larger spatial dimensions, say, $3 \times 3$ or $5 \times 5$, do yield better performance \cite{DBLP:journals/corr/SimonyanZ14a}.  If similar objects are present in different regions in an image, kernels of higher spatial sizes have a higher potential in detecting these similarities.  We will do a comparative study on kernels of various spatial sizes.

\vspace*{-5mm}
\section{Interpretable Saak Features} \label{sec:Interpretable Saak Features}

In this section, we describe the developed techniques used for analyzing, learning, and extracting multi-stage Saak features. Visualization is employed to analyze and have an understanding of these features to learn the optimal ones for required application. New strategies and algorithms based on the feature analysis are suggested for selection and extraction of the optimal feature representations from Saak transform. 

\subsection{Visualization} \label{subsec:Visualization}

Visualizing features is to gain intuition about the network model. We visualize the feature representation of Saak transform at different stages. Figure \ref{fig:Figure-1} shows examples of Saak features extracted from STL-10 dataset. Three random images from each class are shown along with their first stage feature responses of Saak transform. It can be seen that the feature responses are stronger in the most important regions of the images. This observation is consistent for most of the images across classes. Furthermore, the important regions of the images captured by Saak features are also retained in the successive stages of Saak transform, revealing different structures that excite those regions. It is important to note that Saak transform is computed in a feed-forward manner without any supervision and back propagation and the whole feature extraction process is completely transparent. 

\subsection{Discriminability of Saak features}\label{subsec:Discriminability of Saak features}

In this section we discuss the discriminability in Saak features. We show that Saak features are able to capture and encode the unique structures across different images and classes. If a certain Saak kernel gives high response to  important locations in an image and if it is consistent across all images of that same class, the coefficients corresponding to that kernel can be considered as a discriminant feature. To demonstrate the discriminant capability, Figure \ref{fig:Figure-1} shows the Saak feature responses to different objects using carefully selected kernels. Specific kernels are chosen for a particular class and they are convoluted with the images from those classes for visualization. We show that the chosen filters cater high responses in certain locations of images, helping us discriminate one class from other classes.

As mentioned above, Figure \ref{fig:Figure-1} shows features for images from STL-10 dataset. To further explain, consider images from bird, deer and horse classes. We carefully choose a kernel that is consistent across all images of these classes. Thus, a particular kernel for deer class gives high response at locations of deer's ears and horns. Similarly, a unique kernel for horse class gives high response for it's legs. Two different kernel outputs are shown for bird class and we can see that, one corresponds to feathers and another corresponds to beak. From these feature visualization, we can recognize that these responses are discriminant. Some important features for airplane are it's body and wings; for ship are it's stern and mast; for cat is it's face; for car is it's wheels; and for monkey, dog and truck is their body itself. This demonstrates the discriminant power of Saak features in both classifying the image and localizing class-specific image regions in a single forward pass. 

Discriminant Saak kernels that work best for one class, give a poor response for other classes. Saak features are not shared by many classes and hence can be used for many computer vision applications. Careful feature selection techniques can help us choose the right set of features for a desired application. We discuss Saak kernels and degrees of freedom in extracting them in section \ref{subsec:Saak_kernels}. Since our main focus for this paper is classification, we propose cross-entropy measure to automatically select discriminant features in section \ref{subsec:Feature_Selection}. 

\subsection{Saak Kernels}\label{subsec:Saak_kernels}

At stage 1, Local Cuboids (LCs) of size $k_{s} \times k_{s} \times K_{0}$ are extracted from input \textbf{f}, where $K_0$ = 1 for monochrome images and $K_0$ = 3 for color images. Conducting KLT transform on these patches yields a set of signed KLT coefficients or Saak coefficients. Spectral dimension of consecutive stages of Saak transform is given by, 
\begin{equation}\label{eqn:multistagekernelsize}
\begin{split}
    & K_p = k_s \times k_s \times 2 \times K_{p-1} \\
    & where \hspace{3mm} p = stage\hspace{1mm}number = 2,3,...
\end{split}    
\end{equation}

The spectral dimension is doubled due to kernel augmentation, as mentioned in the third term of RHS of (\ref{eqn:multistagekernelsize}). With this recursive computation of multistage Saak LCs, there is an exponential growth of spectral dimensions with respect to stage number $p$ (\ref{eqn:kernelsizegrowth}). 
\begin{equation}\label{eqn:kernelsizegrowth}
    K_p = (k_s \times k_s \times 2)^p
\end{equation}

Reduction in spatial dimension is inversely proportional to the kernel size of non-overlapping LCs extracted. Degrees of freedom for Saak kernel extraction lie in selection of 1) $k_s$ (kernel size), 2) overlapping or non-overlapping LCs and 3) stopping criterion (final stage spatial dimension). 

Some datasets are simpler and more structured like MNIST, while other datasets contain complex objects with diverse background like CIFAR-10, STL-10 and ImageNet. Saak kernels of different sizes needs to studied and modified depending on the variations in the images. The $2 \times 2$ cuboids extracted from MNIST dataset are sufficient for classification since variations within an image are less. Stack of $3 \times 3$ LCs \cite{DBLP:journals/corr/SimonyanZ14a}, result in better performance when compared to other kernel dimensions during convolutions. When similar objects are present in different regions of the image, LCs with higher spatial dimension have higher potential to detect these similarities. Hence, we experiment Saak kernels of increased sizes like $3 \times 3$ and $5 \times 5$. 

We exploit spatial redundancy in the architecture using overlapping kernels. This increases robustness of the model against changes in the input. Overlapping cuboids with max-pooling layer reduces spatial dimensions in both vertical and horizontal directions. Some advantages of max-pooling are that it is completely data dependent and increases non-linearity in the model. 

Multi-stage Saak transform can be understood as recursive decomposition of an image into four quadrants to form a quad-tree structure with its root being the image itself and leaf being LCs. The stopping criterion for last stage can be when one set of signed KLT coefficients of dimension $1 \times 1 \times K_{f}$ is obtained. For an image of size $2^{P} \times 2^{P}$, $K_{f} = 2^{3P}$. Final stage responses of dimension $1 \times 1$ works well for simple images like MNIST, but in most cases, they are not be able to capture complex structure in the images. For these cases, final stage has high cross-entropy when compared to the previous stages. Accordingly, multi-stage Saak transform can be stopped at early stages.

\subsection{Feature Selection}\label{subsec:Feature_Selection}

Input \textbf{f} is convoluted with extracted Saak kernels to extract Saak features. Based on kernel size $k_{s}$, spatial resolution of feature responses reduce at every stage. Consider block of feature responses at any stage $p$ for a single image as {\boldmath${f_{p}}$} with dimension $D_{p1} \times D_{p2} \times K_p$. First two dimensions represent spatial dimension along vertical and horizontal directions. The third one represents the spectral dimension of feature responses for an image. If there are $N$ images in the training data, then total dimension of feature responses can be given as $N \times D_{p1} \times D_{p2} \times K_p$. 

Cross-entropy for feature responses is calculated at every index $(i,j,k)$, where $(i,j)$ represents spatial location and $k$ represents spectral dimension. Let $C$ be the number of classes. Entropy at every location is given by,

\begin{equation}\label{eqn:cross entropy}
    H = \sum_{n=1}^{N}\sum_{c=1}^{C}{y_{n,c}\hspace{0.5mm}\log{\frac{1}{p_{n,c}}}}
\end{equation}
where
\begin{equation}\label{eqn:ync}
   y_{n,c} = 
   \begin{cases}
            1, &  \text{if } \boldsymbol{f_p}(n,i,j,k) \in c\\
            0, &  \text{if } \boldsymbol{f_p}(n,i,j,k) \notin c
    \end{cases}
\end{equation}
and $p_{n,c}$ is the probability of $n^{th}$ sample in class $c$. To obtain this, feature response values at $(i,j,k)$ location across all images are taken. Histogram of these $N$ values is calculated using a certain number of bins, $B$. From various experiments, we concluded that feature selection is stable irrespective of number of bins. We choose $B =10$ and proceed by getting
\begin{equation}\label{eqn:bins entropy}
    mc = (mc_1, mc_2, ..., mc_B),
\end{equation}
where $mc_i$ represents maximum occurring class in bin $i$, and $mc_i
\in {1,2,...,C}$. Probability $p_{n,c}$ is determined as
\begin{equation}\label{eqn:pnc}
    p_{n,c} = \frac{\sum_{i=1}^{B}\mathds{1}(mc_i = c)}{B}
\end{equation}

At the end, $D_{p1} \times D_{p2} \times K_{p}$ cross-entropy values will be computed at stage $p$. Lower the entropy value at a location, higher is the discriminant power. For every spectral dimension, $D_{p1} \times D_{p2}$ pixels are ranked according to their entropy. First few pixels with lowest cross-entropy values are retained, and others are made zero. This localizes salient regions in an image. Similarly, spatially averaged cross-entropy for all spectral dimensions is
obtained. From these average values, spectral dimensions are ranked, and first few $K_p^{'}$ with lowest average cross-entropy values are chosen. Thus spatially sparse feature responses with dimension $D_{p1} \times D_{p2} \times K_p^{'}$ are chosen at stage p. This is repeated at all stages of multi-stage Saak transform for classification. 

%%%%%%%%%%%%%%%%%%%%%%%%%%%%%%%%%%%%%%%%%%%%%%%%%
\begin{figure}
\begin{center}
\includegraphics[width=1\linewidth]{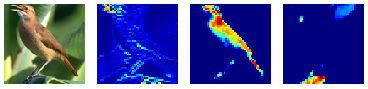}
\end{center}
\vspace*{-6mm}
\caption{Three different Saak feature maps for a bird image. The maps capture different regions of information. First two images contain discriminant regions (beak and body) whereas the third (leaf) doesn't.}
\label{fig:Figure-2}
\end{figure}
%%%%%%%%%%%%%%%%%%%%%%%%%%%%%%%%%%%%%%%%%%%%%%%%%

To understand the motivation behind the cross-entropy based feature selection, we use feature visualization in Figure \ref{fig:Figure-2}, which shows three different feature maps corresponding to a bird image. First feature map has a high response for the bird's beak, second map has a high response for the bird's body (mainly its feathers), and the last map has a high response for the leaf. The cross-entropy value for the last feature map is high when compared to the first and second maps. Lower the cross-entropy, higher is the discriminant information for that particular class. 

%%%%%%%%%%%%%%%%%%%%%%%%%%%%%%%%%%%%%%%%%%%%%%%%%
\begin{figure}
    \centering
    \begin{subfigure}[b]{0.45\textwidth}
        \includegraphics[width=1\textwidth]{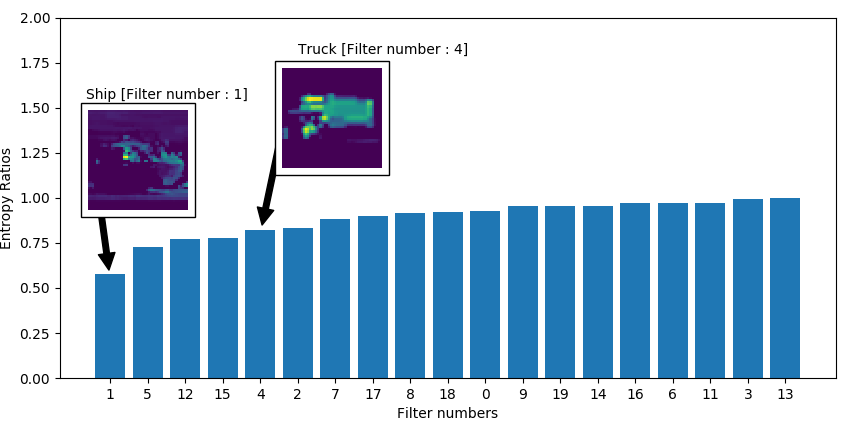}
	\caption{filter numbers vs cross-entropy}
        \label{fig:Figure-3a}
    \end{subfigure}
    ~ %add desired spacing between images, e. g. ~, \quad, \qquad, \hfill etc. 
    \begin{subfigure}[b]{0.45\textwidth}
        \includegraphics[width=1\textwidth]{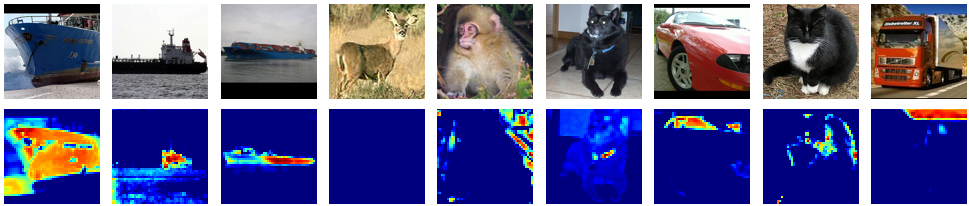}
        \caption{Sample responses for lowest cross-entropy filter}
        \label{fig:Figure-3b}
    \end{subfigure}
 \vspace*{-2.5mm}
\caption{ (a) shows relationship between Saak filters and the computed cross-entropy values for the first stage; (b) responses of filter with lowest cross-entropy on class 'ship' vs other classes}\label{fig:Figure-3}
\end{figure}

%%%%%%%%%%%%%%%%%%%%%%%%%%%%%%%%%%%%%%%%%%%%%%%%%

Figure \ref{fig:Figure-3} (a) shows the relationship between Saak transform filters and the computed cross-entropy values for the filters. The first kernel with the lowest cross-entropy gives discriminant features for ship class. Similarly fourth kernel gives a discriminant feature for the truck class. For instance, when the first kernel is convoluted with all images, it highlights the important features of ship (stern and mast) when the image is ship, whereas gives undesired responses when the image is not a ship. When the fourth kernel is convoluted with an image of a truck, resulting feature response highlights the truck's body, but a similar response is not obtained from other classes. We can interpret this from Figure \ref{fig:Figure-3} (b), the first three images correspond to the responses obtained by the lowest cross-entropy filter on 'Ship' class and the other images contain the responses of the same filter on other classes.

%%%%%%%%%%%%%%%%%%%%%%%%%%%%%%%%%%%%%%%%%%%%%%%%%
\begin{figure*}
\begin{center}
\includegraphics[width=1\linewidth]{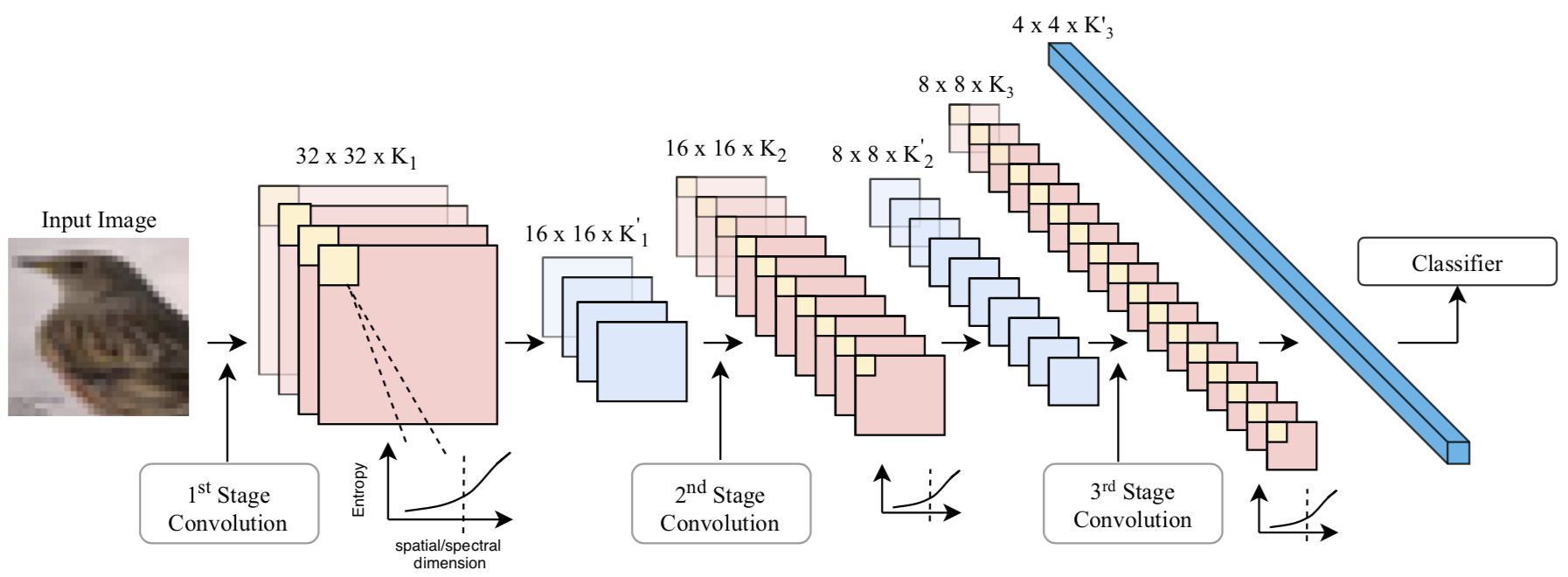}
\end{center}
\caption{3-stage Saak architecture for CIFAR-10. Input image is of size $32 \times 32 \times 3$. Spatial dimension reduces every stage and final stage has response of size $4 \times 4  \times K_{3}^{'}$. This is reshaped and fed as input to the classifier.}
\vspace*{-5mm}
\label{fig:block diagram}
\end{figure*}
%%%%%%%%%%%%%%%%%%%%%%%%%%%%%%%%%%%%%%%%%%%%%%%%%

\section{Applications} \label{sec:Applications}

Saak transform offers great potential to many computer vision and machine learning tasks. For concreteness, we focus on two applications in this paper, i.e. image classification and adversarial defense to demonstrate the unique capabilities of Saak features learnt from images, though the technique is applicable to other forms of data as well. 

\subsection{Image Classification}\label{subsec:Image Classification}

We revisit image classification problem using Saak transform theory. There is major difference between the deep neural network and the Saak transform methodology. For example, deep neural network converts image classification problem to multiple object segmentation problems using object proposals. Saak transform does not need any bounding boxes. As stated before, Saak transform offers a family of joint spatial-spectral representations that have capabilities for both classifying images and localizing class-specific image regions. It tackles the challenging segmentation task directly based on the spatial-spectral information. After the segmentation task, it provides a semantic label to each region. 

Saak transform consists of 1) Extracting LCs from the images, 2) Obtaining KLT components, 3) Convoluting the images with the extracted kernels, 4) Calculating the cross entropy measures, 5) Selecting the best spatial/spectral components. Our entire approach is shown in the Figure ~\ref{fig:block diagram}. Images are convoluted with all the kernels obtained. The best coefficients are then chosen based on the lowest cross-entropy values calculated. This is done on both spectral and spatial dimensions. The responses corresponding to the best kernels are then fed as inputs to the next stage of Saak transform. The architecture in Figure \ref{fig:block diagram}, shows multi-stage Saak transform using $3 \times 3$ Saak kernels, applied to CIFAR-10 images. After 3 stages of Saak transform, we end up with a spatial dimension of $4 \times 4$ and a spectral dimension $K_3$. These responses are then reshaped and used for classification. 

\subsection{Defense Against Adversarial Attacks}\label{subsec:Defense Against Adversarial Attacks}

Effective defense against adversarial attacks has been of great concern in designing machine learning-based vision systems. It has been shown that deep neural network is vulnerable to adversarial attacks \cite{deep_adversarial_1}. These attacks come in the form of adversarial inputs with carefully crafted imperceptible perturbations added to the input images, which can drastically cause learning systems to misinterpret adversarial images.

Methods to defend adversarial attacks have been done through adversarial training, adversarial detection, gradient masking methods, etc \cite{Tramr2017EnsembleAT, Miyato2018VirtualAT, DBLP:journals/corr/LuIF17, DBLP:journals/corr/GrosseMP0M17}. Adversarial training becomes specific to attack methods and fail to generalize while adversarial detectors still possess the risk of being fooled by the attacker. We show how our Saak feature based method is robust to such small perturbations in an image. 

\begin{figure}
\begin{center}
\includegraphics[width=1\linewidth]{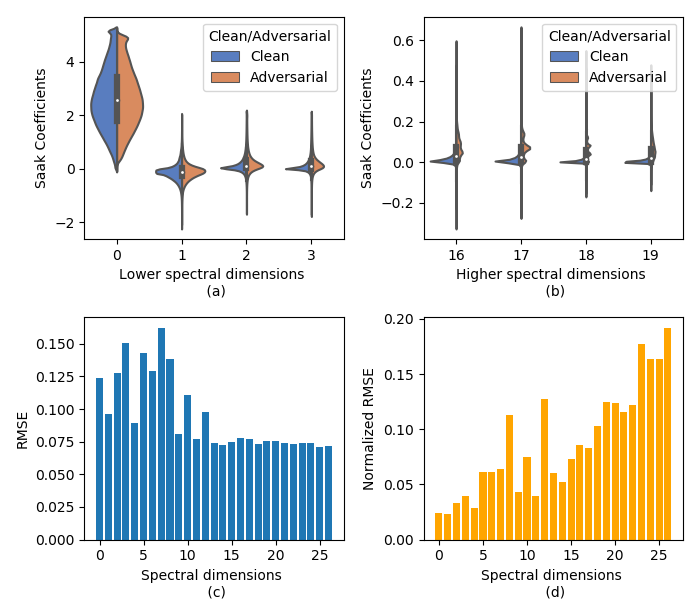}
\end{center}
\vspace*{-6 mm}
\caption{(a) and (b) show distribution of Saak coefficients for lower and higher spectral dimensions respectively, (c) and (d) are the RMSE and normalized RMSE between clean and FGSM attacked Saak coefficients in different spectral dimensions}
\label{fig:adversarial_analysis_plot}
\end{figure}

In the Saak transformation domain, clean and adversarial images have different distributions at different spectral dimensions. Careful selection of the spectral dimensions at every stage, can be viewed as an automatic noise filtering technique. Figure \ref{fig:adversarial_analysis_plot} shows distribution of Saak components belonging to first few spectral dimensions, followed by the distribution for higher spectral dimensions. Saak spectral components differ for both clean and adversarial images at higher dimension. We also show the normalized and the original RMSE (root-mean-squared-error) values between clean and FGSM adversarial samples in different spectral components. We can observe from plots (c) and (d) that clean and adversarial samples have different Saak coefficient values in high spectral dimensions. These results were obtained from first stage Saak transform of CIFAR10 images using $3 \times 3$ local cuboids.

We classify adversarial images using Saak transform. We extract kernels using clean images and follow the same procedure as we classify clean images. As shown in Figure \ref{fig:flow chart} Saak kernels are used to extract the coefficients from attacked images. We classify adversarial attacked images after selecting features using our cross-entropy based method.

%%%%%%%%%%%%%%%%%%%%%%%%%%%%%%%%%%%%%%%%%%%%%%%%%
%%%%%%%%%%%%%%%%%%%%%%%%%%%%%%%%%%%%%%%%%%%%%%%%%

\section{Experiments}\label{sec:experiments}

We extensively conduct experiments aimed at deeply understanding the Saak feature representation and demonstrating its benefits and utilities to computer vision tasks.
We study classification performance under different settings to provide in-depth examinations on the unique properties of Saak features in terms of discriminant power, robustness, and complexity.

Three well-known datasets are used for experiments: MNIST, CIFAR-10 and STL-10. The MNIST \cite{MNIST} dataset contains 60,000 training images and 10,000 testing images of handwritten digits.  CIFAR-10 \cite{CIFAR-10} dataset contains 60,000 32x32 color images in 10 different classes.  STL-10 \cite{STL-10} dataset is inspired from CIFAR10 dataset but contains higher resolution images (96x96) that make it a challenging benchmark for developing more scalable algorithms.

\subsection{Effect of Overlapping Local Cuboids} \label{subsec:Effect of Overlapping Local Cuboids}

In the original Saak architecture given in \cite{DBLP:journals/corr/abs-1710-10714}, non-overlapping Local Cuboids (LCs) were extracted. In the new architecture, we use overlapping kernels to maximize the information contained in each LC. However, this will result in an exponential increase in the number of features at each Saak stage. To reduce the dimension, we employ max-pooling for sub-sampling and construct multi-stage Saak transform. In addition, it has been proven that max-pooling is efficient in capturing variances in images after convolution with stride one \cite{DBLP:Max-pooling}. 

Table \ref{table:Max-pooling} shows the classification performance using overlapping cuboids and non-overlapping cuboids. As in the case of CNNs, max-pooling increases the performance, since it increases non-linearity in the model. There is increased redundancy in the overlapping LCs extraction, leading to better KLT components. Thus, overlapping cuboids with max-pooling has enhanced the classification performance. For example, the accuracy for MNIST is $99.3\%$, for CIFAR-10 and STL-10, the accuracy has increased by $13.3\%$ and $8.75\%$ respectively.

%%%%%%%%%%%%%%%%%%%%%%%%%%%%%%%%%%%%%%%%%%%%%%%%%
\begin{table}
\begin{center}
\begin{tabular}{c c c}
\hline
\multirow{2}{*}{Dataset} & \multicolumn{2}{c}{Accuracy}  \\
\cline{2-3}
   &  Non-Overlapping & Overlapping \\
\hline \hline
MNIST & 98.81\%  & \textbf{99.3\%} \\
CIFAR-10 & 61.3\% & \textbf{74.6\%} \\
STL-10 &  54.3\% & \textbf{63.05\%} \\
\hline
\end{tabular}
\end{center}
\vspace{-5mm}
\caption{Effect of overlapping Local Cuboids and max-pooling to classification performance}
\label{table:Max-pooling}
\end{table} 
%%%%%%%%%%%%%%%%%%%%%%%%%%%%%%%%%%%%%%%%%%%%%%%%%

\subsection{Effect of Kernel Dimension}\label{subsec:Effect of Kernel Dimension}

We evaluate the impact of Saak kernel dimension on the classification performance.  We experiment different kernel dimensions of $2 \times 2$, $3 \times 3$ and $5 \times 5$ to change the receptive fields. Table \ref{table:Kernel} contains test accuracy obtained for different kernel dimensions for all three datasets.  

Since overlapping kernel extraction gives superior performance, our experiments on kernel dimensions are also based on the same. The results show a good increase in the accuracy when the kernel dimension was increased from $2 \times 2$ to $3 \times 3$. Larger receptive fields help in localization of required objects inside the images. However, the performance does not increase when kernel dimension was changed from $3 \times 3$ to $5 \times 5$. When $5 \times 5$ LCs are extracted, the number of stages of Saak transform reduces to 2, and hence the spatial dimension at the last stage is very high when to compared to 3-stage Saak transform with $3 \times 3$ kernels. This will result in over-fitting due to high feature dimension. 

The change of kernel dimension has provided a significant boost in the performance for CIFAR-10 and STL-10. Larger kernel dimension gives better performance when objects in the images are at different locations. For MNIST, since the digits are mostly located in the center, there is a very small increase in the performance when kernel dimension is increased. In the case of CIFAR-10 and STL-10, the objects are not always present in the center. Thus, an increased receptive field captures the discriminant regions and improves performance. Saak transform’s capability of localizing the discriminative regions in the image is verified.

%%%%%%%%%%%%%%%%%%%%%%%%%%%%%%%%%%%%%%%%%%%%%%%%%
\begin{table}
\begin{center}
\begin{tabular}{c c c c}
\hline
\multirow{2}{*}{Dataset} & \multicolumn{3}{c}{Accuracy}  \\
\cline{2-4}
   &  $2 \times 2$ & $3 \times 3$ &  $5 \times 5$\\
\hline \hline
MNIST & \textbf{99.30}\% & 98.94\% & 99.03\% \\
CIFAR-10 & 65.68\% & \textbf{74.6\%} & 73.06\% \\
STL-10 & 55.30\% &  \textbf{63.05\%} & 59.50\% \\
\hline
\end{tabular}
\end{center}
\vspace{-5mm}
\caption{Effect of kernel dimension to classification performance}
\label{table:Kernel}
\end{table} 
%%%%%%%%%%%%%%%%%%%%%%%%%%%%%%%%%%%%%%%%%%%%%%%%%

\subsection{Robustness}\label{subsec:Robustness}

To study the robustness of Saak features, we conduct experiments using images attacked by state of the art attack methods:  Deepfool \cite{DBLP:journals/corr/Moosavi-Dezfooli15}, FGSM \cite{FGSM} and BIM \cite{BIM}. Adversarial images are created by applying a small perturbation to an image in a way that changes the predictions made by a pre-trained model. Classification accuracy results for attacked images are shown in Table \ref{table:adversarial} for different datasets. Many deep learning methods are vulnerable to adversarial images \cite{DBLP:journals/corr/SzegedyZSBEGF13}. Unlike other methods, Saak transform based classification is robust and the accuracies of clean and attacked images are very close.

We flexibly combine different classifiers with the extracted Saak features intending to evaluate the overall performance of a complete classification system. Four classifiers are tested, i.e. Random Forest (RF), Support Vector Machines (SVM), Logistic Regression (LR) and Multi-layer Perceptron (MLP), though the technique is applicable to any classifier that accepts feature representations.  Table \ref{table:Classification} shows the test accuracy corresponding to each classifier.

%%%%%%%%%%%%%%%%%%%%%%%%%%%%%%%%%%%%%%%%%%%%%%%%%
\begin{table}
\begin{center}
\begin{tabular}{c c c c c}
\hline
Attack & FGSM & BIM & DeepFool\\
\hline\hline
MNIST &  94.52\%& 94.13\%& 95.51\% \\
CIFAR-10 &  49.50\%& 48.50\%& 70.44\%\\
STL-10& 47.69\%&  50.55\%& 58.5\%\\
\hline
\end{tabular}
\end{center}
\vspace{-5mm}
\caption{Performance comparison on different adversarial attacks}
\label{table:adversarial}
\end{table} 

%%%%%%%%%%%%%%%%%%%%%%%%%%%%%%%%%%%%%%%%%%%%%%%%%

%%%%%%%%%%%%%%%%%%%%%%%%%%%%%%%%%%%%%%%%%%%%%%%%%
\begin{table}
\begin{center}
\begin{tabular}{c c c c c}
\hline
\multirow{2}{*}{Dataset} & \multicolumn{4}{c}{Accuracy}  \\
\cline{2-5}
    & LR & RF & SVM& MLP\\
\hline \hline
MNIST & 98.77\% & 96.90\% & 98.5\%& \textbf{99.30\%} \\
CIFAR-10 &64.43\% & 55.76\% & 58.5\%& \textbf{74.60\%} \\
STL-10 & 54.70\% & 45.90\% & 49.43\%& \textbf{63.05\%} \\
\hline
\end{tabular}
\end{center}
\vspace{-5mm}
\caption{Performance comparison on different classifiers}
\label{table:Classification}
\end{table} 

%%%%%%%%%%%%%%%%%%%%%%%%%%%%%%%%%%%%%%%%%%%%%%%%%

\subsection{Complexity}\label{subsec:Complexity}

Complexity analysis is gaining popularity in developing practical machine learning frameworks. The main aim is to understand if added complexity is worth the benefits. In similar lines, we conduct complexity analysis of Saak transform based feature representation. KLT components of different stage Saak kernels are considered for this experiment. Images are selected on the basis of stratified sampling. Table \ref{table:cosine similarity} displays cosine similarity of the components between subset of images and the whole dataset.  We experiment with different dataset sizes like 5000, 10000, 20000, 30000 and 40000 and compare with 50000 images from CIFAR-10. We can see that the components extracted at each stage with different number of images are very stable. The KLT components from smaller dataset sizes are very similar to the components extracted from the entire training set. This is very advantageous in constrained environments where huge number of training images cannot be used. 

\begin{figure}
\begin{center}
\includegraphics[width=1\linewidth]{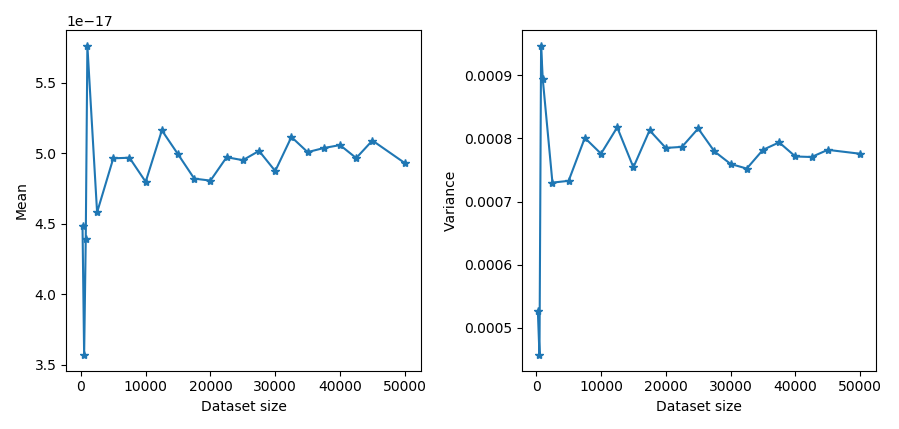}
\end{center}
\vspace*{-7mm}
\caption{Mean and variance of last stage KLT components obtained using different subsets CIFAR-10 training images}
\label{fig:complexity}
\end{figure}

Mean of the KLT components against number of images used for the extraction is shown in Figure \ref{fig:complexity} (a). Similarly Figure \ref{fig:complexity} (b) shows variance of KLT components. A convergence in mean and variance can be observed with use of just 25\% (20000 samples) of the training dataset. Hence, number of images needed for kernel extraction plays a less significant role in performance, thereby greatly reducing the complexity of the whole system.

%%%%%%%%%%%%%%%%%%%%%%%%%%%%%%%%%%%%%%%%%%%%%%%%%
\begin{table}
\begin{center}
\begin{tabular}{c c c c}
\hline
Size & Stage 1 & Stage 2 & Stage 3 \\
\hline\hline
5000 &  0.9999 & 0.9885 & 0.9632 \\
10000 & 0.9999 & 0.9877 & 0.9851 \\
20000 & 0.9999 & 0.9881 & 0.98492 \\
30000 & 0.9999 & 0.9914 & 0.9816 \\ 
40000 & 0.9999 & 0.9999 & 0.9999 \\
\hline
\end{tabular}
\end{center}
\vspace{-5mm}
\caption{Cosine Similarity between KLT coefficients obtained using different subsets of images and the entire CIFAR-10 dataset}
\label{table:cosine similarity}
\end{table} 

%%%%%%%%%%%%%%%%%%%%%%%%%%%%%%%%%%%%%%%%%%%%%%%%%

%%%%%%%%%%%%%%%%%%%%%%%%%%%%%%%%%%%%%%%%%%%%%%%%%
%%%%%%%%%%%%%%%%%%%%%%%%%%%%%%%%%%%%%%%%%%%%%%%%%

%%%%%%%%%%%%%%%%%%%%%%%%%%%%%%%%%%%%%%%%%%%%%%%%%

\section{Conclusion and Future Work } \label{sec:Conclusion and Future Work}

This paper studies the power of Saak features as an effort towards interpretable deep learning. We have provided an in-depth examination on the unique properties of Saak features in terms of discriminant power, robustness, and complexity.  Multiple Saak transforms are generalized on top of the basic Saak transform to transfer image to a joint spatial-spectral feature representation. Visualization approach is developed to analyze and understand Saak representation to learn the optimal ones for required applications. New strategies and algorithms based on the feature analysis are proposed for selection and extraction of the optimal feature representation. Two applications with extensive experiments are developed to demonstrate the benefits and utilities of Saak transform and features. 

Saak representation is entirely a new signal transform concept. On one hand, it is a complete data-driven transform that provides an unsupervised one-pass mechanism. On the other hand, its complexity is much lower than that of state-of-the-art and it can be effectively implemented in portable computing devices. Most importantly, Saak process is completely transparent and interpretable with solid theoretical support. We will continue to shed light on its theory and applications.

\section{Acknowledgments}
This research was supported by the U.S. Army Research Laboratory's External Collaboration Initiative (ECI) of the Director's Research Initiative (DRIA) program. The views and conclusions contained in this document are those of the authors and should not be interpreted as representing the official policies, either expressed or implied, of the U.S. Army Research Laboratory or the U.S. Government. The U.S. Governments are authorized to reproduce and distribute reprints for Government
purposes notwithstanding any copyright notation hereon.

{\small
\bibliographystyle{ieee}
\bibliography{egbib}
}

\end{document}